\documentclass{article}
\usepackage{spconf,amsmath,graphicx,hyperref}

\usepackage{cite}
\usepackage{amsmath,amssymb,amsfonts}
\usepackage{graphicx}
\usepackage{textcomp}
\usepackage{xcolor}
\usepackage{symbolDef}
\usepackage{subcaption}
\usepackage{theorem}
\def\BibTeX{{\rm B\kern-.05em{\sc i\kern-.025em b}\kern-.08em
    T\kern-.1667em\lower.7ex\hbox{E}\kern-.125emX}}

\newtheorem{theorem}{\bf Theorem}

\def \QED {$\blacksquare$}
\usepackage{hyperref}
\usepackage{booktabs}
\usepackage{multirow}
\usepackage{algorithm}
\usepackage{algpseudocode}

\ninept

\algrenewcommand\algorithmicend{\textbf{end}}

\makeatletter
\algdef{SE}[FOR]{For}{EndFor}[1]{\algorithmicfor\ #1\ \algorithmicdo}{\algorithmicend}
\makeatother

\input{mysymbols.sty}

\title{
Precision Neural Networks: Joint Graph and Relational Learning
}
\name{Andrea Cavallo$^*$, Samuel Rey$^\dagger$, Antonio G. Marques$^\dagger$, Elvin Isufi$^*$
\thanks{
Work partially funded by the TU Delft AI Labs programme, the NWO OTP GraSPA proposal \#19497, the NWO VENI proposal 222.032, the Spanish AEI (10.13039/501100011033) grant PID2022-136887NB-I00, and the Community of Madrid via the Ellis Madrid Unit and grants URJC/CAM F1180 and TEC-2024/COM-89.
Emails:  
    \{a.cavallo, e.isufi-1\}@tudelft.nl, 
    \{samuel.rey.escudero, antonio.garcia.marques\}@urjc.es.
}}
\address{
$^*$Delft University of Technology, Delft, Netherlands \\
$^\dagger$King Juan Carlos University, Madrid, Spain
}
\begin{document}
%
\maketitle
\begin{abstract}
CoVariance Neural Networks (VNNs) perform convolutions on the graph determined by the covariance matrix of the data, which enables expressive and stable covariance-based learning. However, covariance matrices are typically dense, fail to encode conditional independence, and are often precomputed in a task-agnostic way, which may hinder performance. To overcome these limitations, we study \emph{Precision Neural Networks (PNNs)}, i.e., VNNs on the precision matrix---the inverse covariance. The precision matrix naturally encodes statistical independence, often exhibits sparsity, and preserves the covariance spectral structure. To make precision estimation task-aware, we formulate an optimization problem that \emph{jointly learns the network parameters and the precision matrix}, and solve it via alternating optimization, by sequentially updating the network weights and the precision estimate. We theoretically bound the distance between the estimated and true precision matrices at each iteration, and demonstrate the effectiveness of joint estimation compared to two-step approaches on synthetic and real-world data.
\end{abstract}

\section{Introduction}
\begin{figure}
    \centering
    \includegraphics[width=\linewidth]{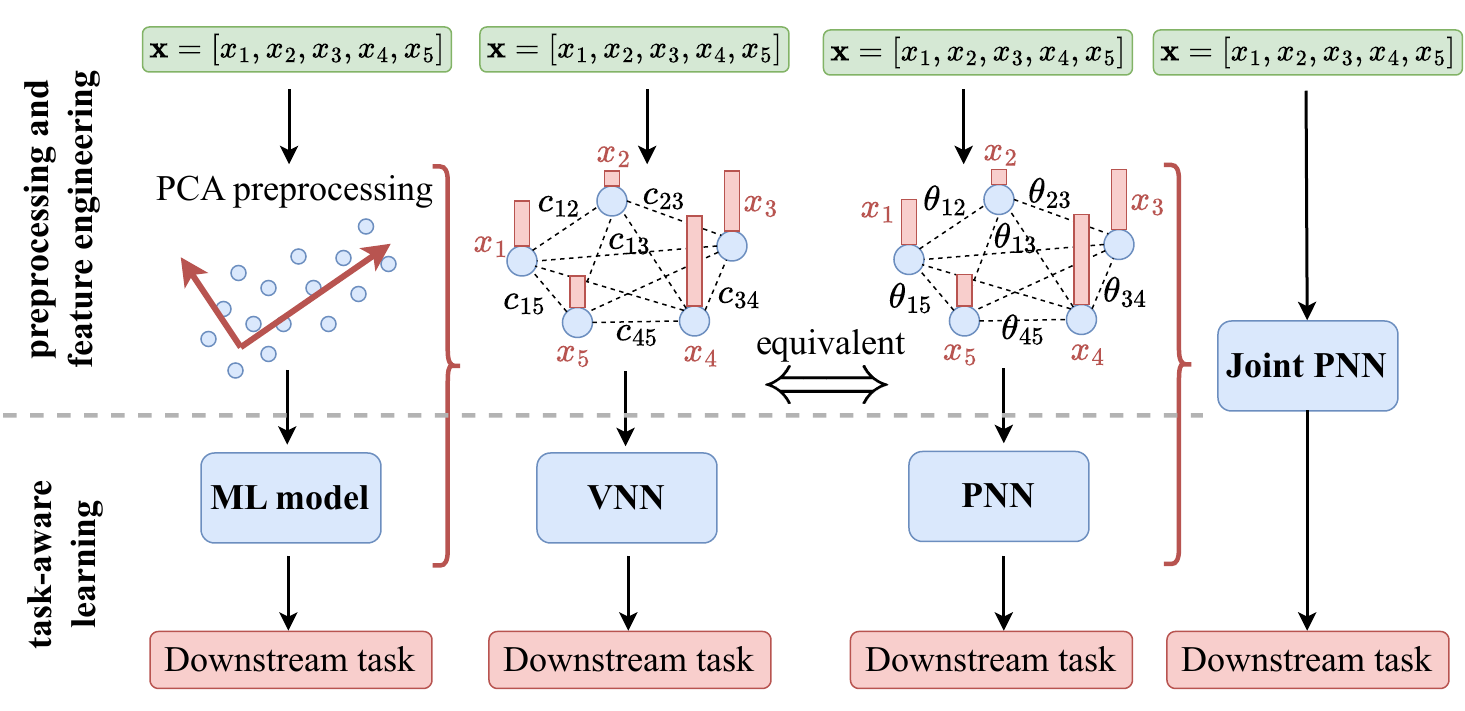}
    \caption{
    Data processing via covariance information. Traditional pipelines use PCA for preprocessing followed by an ML model for the downstream task. VNNs and PNNs merge these steps by learning end-to-end on the covariance or the precision graph. Joint PNNs unify the estimation of the precision matrix and the network weights. 
    \vspace{-.8cm}
    }
    \label{fig:method}
\end{figure}

Capturing relations among data points in complex systems is fundamental to their effective modeling and understanding.
In applications such as brain data~\cite{li2021braingnn,bessadok2022graph}, sensor measurements~\cite{liao2022har}, and finance~\cite{palomar2024portfolio,cardoso2020algorithms}, data interactions are commonly captured via covariance-based processing. 
A classical approach is Principal Component Analysis (PCA)~\cite{Jolliffe2002pca}, which projects the data onto the eigenvectors of the covariance matrix to maximize variance and, when desired, reduce dimensionality. 
While effective as a preprocessing step before applying a machine learning (ML) model for a downstream task (cf. Figure~\ref{fig:method}), PCA is sensitive to noise and estimation errors, and lacks flexibility in processing the covariance eigenvectors. 
To overcome these limitations, coVariance Neural Networks (VNNs)~\cite{sihag2022covariance} have emerged as powerful tools thanks to their stability to finite sample noise, expressivity, and ability to learn directly from data correlations. 
VNNs are sequences of nonlinearities interleaved with banks of covariance filters, i.e., polynomials of the covariance matrix, which can be interpreted as convolutions on the graph determined by the covariance matrix. 
This enables VNNs to learn expressive functions in the covariance eigenspace, thus extending PCA towards covariance-aware learning for downstream tasks (cf. Figure~\ref{fig:method}), while being stable to covariance estimation errors. This makes VNNs robust in low-data regimes, favorable for interpretability~\cite{sihag2024explainable} and transferable across resolutions~\cite{sihag2024transferability}, and these benefits extend to spatiotemporal data~\cite{cavallo2024stvnn} and to covariance estimators promoting sparsity~\cite{cavallo2024sparse} and fairness~\cite{cavallo2024fairvnn}. 
However, VNNs and their extensions still rely on precomputed covariances, which decouples correlation estimation from the downstream objective (e.g., brain connectivity in healthy versus diseased subjects, or horizon-specific financial dependencies). 
Moreover, the covariance matrix only reflects marginal correlations and cannot distinguish direct from indirect effects, making it insufficient for capturing conditional independence and often leading to redundant dependency structures.

As an alternative, we study the application of VNNs to the precision matrix, thus defining \emph{Precision Neural Networks (PNNs)}, and we propose a joint estimation of the precision and the neural network (NN) weights.
The precision matrix is typically sparse and encodes conditional independence among variables, thereby capturing covariance information in a more efficient and sparse representation and more effectively modeling data relations~\cite{friedman1991}. 
Moreover, it shares the eigenvectors of the covariance matrix, which enables PNNs to perform the same spectral task as VNNs.
However, directly inverting the sample covariance is unreliable in high-dimensional regimes, and existing sparse precision estimators such as graphical lasso~\cite{lauritzen1996graphical,friedman2007lasso,rey2023enhanced} are task-agnostic.
To overcome this limitation, we introduce a framework that jointly learns both the sparse precision matrix and the weights of the PNN, yielding a \emph{task-aware precision} that balances data fidelity with downstream performance and can identify connectivity patterns that are relevant to solving the task at hand.
Since the resulting problem is challenging due to the multiple optimization variables, we address it through an alternating strategy that sequentially updates the network parameters and the precision. We provide theoretical insights by proving that, at each iteration, the estimated precision converges to the true matrix at a rate inversely proportional to the square root of the number of observations. 
While several works analyze joint estimation of a graph structure and a model to process it~\cite{jin2020graph,rey2023robust,liu2025two,egilmez2018graph,natali2020topology}, they either focus on graph filters only, they lack a theoretical analysis of the estimated graph or they do not build a connection with covariance-based learning methods.

We summarize our main contributions below.

\noindent \textbf{(C1)} We study PNNs, graph convolutional NNs on sparse precision matrices that leverage conditional independence through an efficient representation.  

\noindent \textbf{(C2)} We develop a joint estimation framework for the precision matrix and the PNN weights, yielding a task-aware precision matrix that adapts to the downstream objective.  

\noindent \textbf{(C3)} We provide theoretical insights on the quality of the estimated precision at each iteration, and demonstrate that PNNs outperform competitive methods on synthetic data and on an age prediction task from cortical thickness measures.

\section{Precision Neural Networks}

We define PNNs, graph convolutional NNs operating on the inverse covariance, known as the precision matrix. 

\smallskip 
\noindent \textbf{Preliminaries.}
Consider a data matrix $\mtX \in \reals^{N\times T}$ (zero-mean w.l.o.g.) whose columns $\vcx_i \in \reals^N$ are observations of a random variable $\vcx$ with covariance $\mtC_0 = \mathbb{E}[(\vcx - \mathbb{E}[\vcx])(\vcx - \mathbb{E}[\vcx])^\Tr]$. 
Since $\mtC_0$ is generally unknown, it is approximated by the sample estimate $\mtC = \mtX\mtX^\Tr / T$ with eigendecomposition $\mtC = \mtV \mtW \mtV^\Tr$, where the diagonal matrix $\mtW \in \reals^{N \times N}$ collects the eigenvalues and $\mtV \in \reals^{N \times N}$ contains the orthogonal eigenvectors. 
The precision matrix and its estimate are, respectively, $\mtTheta_0=\mtC^{-1}$ and $\mtTheta$.

\smallskip
\noindent \textbf{PNN definition.}
A PNN consists of $L$ layers, each comprising a (graph-based) filter bank of size $F_\textnormal{in}^{(\ell)} \times F_\textnormal{out}^{(\ell)}$ followed by a nonlinear activation function $\sigma$. 
For layer $\ell = 1, \dots, L$ and output feature $f = 1, \dots, F_\textnormal{out}^{(\ell)}$, the precision filter and propagation rule are 
\begin{align}
    &\vcx_f^{(\ell)} = \sigma\left(\sum_{j=1}^{F_\textnormal{in}} \mtH_{fj}^{(\ell)}(\mtTheta)\vcx_j^{(\ell-1)}\right), \label{eq:pnn} \\
    &
    \mtH_{fj}^{(\ell)} (\mtTheta) = \sum_{k=0}^K h_{kfj}^{(\ell)} \mtTheta^k. \label{eq:precision_filter}
\end{align}
%
The PNN output $\vcx^{(L)}$ is passed to a readout function, typically a Multi-Layer Perceptron (MLP), to solve the downstream task. 
We denote the overall pipeline, comprising the PNN and the readout, as $\Phi(\vcx, \mtTheta, \vch)$, where $\vcx$ is an input signal and $\vch$ is a vector stacking all polynomial coefficients $h_{kfj}^{(\ell)}$ in the PNN. 
Given a training set $(\mtX_\textnormal{tr}, \vcy_\textnormal{tr})$, where $\mtX_\textnormal{tr} \in \reals^{N \times T}$ contains the input signals and $\vcy_\textnormal{tr} \in \reals^T$ the corresponding labels or regression targets, 
the weights $\vch$ are learned by solving
\begin{equation}\label{eq:training_pnn}
    \min_\vch \mathcal{L}_\textnormal{task}(\vcy_\textnormal{tr}, \Phi(\mtX_\textnormal{tr},  \mtTheta, \vch)) + \beta\|\vch\|_2^2
\end{equation}
where $\mathcal{L}_\textnormal{task}$ is a task-specific loss (e.g., cross-entropy for classification or MSE for regression), and $\beta \geq 0$ is a regularization parameter.

The use of precision matrices in a deep NN enjoys three key benefits.
First, while the covariance matrix encodes pairwise correlations and is typically dense, $\bbTheta$ often admits a sparse structure, enabling more efficient computations. 
Second, when the observed data $\bbx$ follows a Gaussian distribution, $\mtTheta$ encodes conditional independence, with $[\mtTheta]_{ij} = 0$ if and only if $[\vcx]_i$ and $[\vcx]_j$ are conditionally independent given the rest~\cite{lauritzen1996graphical}. 
Finally, $\mtTheta$ is closely connected to Gaussian graphical models, so that \eqref{eq:precision_filter} can be interpreted as a convolution on the graph induced by the precision matrix.

\smallskip 
\noindent \textbf{Connection to PCA.}
An important property of $\mtTheta$ is that it shares the eigenvectors of $\mtC$, so that $\mtTheta = \mtV \mtM \mtV^\Tr$ with $\mtM = \mtW^{-1}$. 
Following the graph signal processing literature~\cite{Gama_2020}, we define the precision Fourier transform as the projection of a signal $\vcx$ on the eigenvectors of the precision matrix, i.e., $\vctx = \mtV^\Tr \vcx$.
Since $\mtV$ is also the eigenbasis of $\mtC$, this transform coincides with the PCA projection. 
Moreover, in the spectral domain, PNNs learn polynomial functions of the eigenvalues of the precision matrix $\{\mu_i\}_{i=1}^N$.
In detail, the action of a single precision filter on the $i$-th component of the projected signal $\vctx$ is
\begin{equation}\label{eq:freq_resp}
    [\vctx]_i^{(\ell)} = \sum_{k=0}^K h_k^{(\ell)} \mu_i^k [\vctx]_i^{(\ell-1)} = h^{(\ell)}(\mu_i) [\vctx]_i^{(\ell-1)}.
\end{equation}
Thus, PNNs modulate the data projections on their principal components $[\vctx]_i$ via their frequency response $h(\mu)$.

\smallskip 
\noindent \textbf{Connection to VNNs.}
VNNs are graph convolutional NNs using the sample covariance as shift operator---i.e., they replace $\mtTheta$ with $\mtC$ in~\eqref{eq:pnn} and thus they learn a polynomial frequency response analogous to \eqref{eq:freq_resp} on the eigenvalues $w=\mu^{-1}$. Consequently, if the filter degrees $K$ are sufficiently large, VNNs and PNNs can learn the same transformation in the spectral domain by properly adjusting the coefficients $h_k^{(\ell)}$.

\section{Joint estimation of PNN parameters and precision matrix}
PNNs require an estimate of the precision matrix to be available. A simple approach is to compute $\mtTheta$ by inverting the sample covariance $\mtC$, but this is known to yield poor estimates, particularly in high-dimensional or low-sample regimes~\cite{friedman2007lasso}. Alternatively, one may employ established methods such as graphical Lasso, which directly estimate $\mtTheta$ under structural assumptions like sparsity. Although these approaches typically provide more reliable precision matrices, they remain task-agnostic, since the estimation procedure does not incorporate the downstream objective. As a result, the obtained precision matrix may still be suboptimal for the learning problem at hand.
To make the estimation fully task-aware, we consider a \emph{joint estimation of the PNN weights and the precision matrix}. 
Given a training set $(\mtX_\textnormal{tr}, \mty_\textnormal{tr})$, we minimize
\begin{align}
    \min_{\vch,\mtTheta} & \;\; \alpha \mathcal{L}_\textnormal{task}(\vcy_\textnormal{tr}, \Phi(\mtX_\textnormal{tr},  \mtTheta, \vch)) + (1 - \alpha) \ccalL_\textnormal{GL} (\mtX_\textnormal{tr},  \mtTheta) \nonumber \\ \textnormal{s.t.} & \quad \mtTheta \succeq 0, \quad  \|\mtTheta\|_2 \leq M, \label{eq:initial_prob}
\end{align}
where $\alpha \in [0,1]$, $\mathcal{L}_\textnormal{task}$ is the task-specific loss from \eqref{eq:training_pnn}, and $\ccalL_\textnormal{GL}$ is the graphical Lasso objective~\cite{friedman2007lasso}, given by
\begin{equation}
     \ccalL_\textnormal{GL} (\mtX_\textnormal{tr},  \mtTheta) = \operatorname{tr}(\mtC \mtTheta) -\operatorname{logdet}(\mtTheta+\epsilon\mtI) + \lambda \| \mtTheta_\mathcal{\bar{D}} \|_1.
\end{equation}
Here, $\mtTheta_\mathcal{\bar{D}}$ denotes the off-diagonal elements of $\mtTheta$, the $\ell_1$ norm promotes sparsity, while the regularization $\epsilon \mtI$ with $\epsilon \geq 0$ enables learning positive semi-definite precision matrices.
$\mtTheta_0$ is assumed to have bounded eigenvalues (see assumptions in Theorem~\ref{th:precision_convergence}), hence the constraint $\| \mtTheta \|_2 \leq M$.
In practice, an effective $M$ can be obtained by overshooting its value based on the minimum eigenvalue of the sample covariance.
The parameter $\alpha$ balances the influence of the task-driven and model-driven terms, yielding a precision matrix $\mtTheta$ that supports the PNN in solving the downstream task. 
However, the joint optimization over $\mtTheta$ and $\vch$ poses a coupled and more challenging landscape, as it leads to complex and potentially unstable gradients~\cite{jin2020graph}.

\subsection{Relaxed Formulation and Optimization.}
To partially alleviate these issues, we start by adopting the following relaxed formulation:
\begin{align}\label{eq:joint_prob}
    \min_{\vch,\tbTheta}  &\quad \alpha \mathcal{L}_\textnormal{task}(\vcy_\textnormal{tr}, \Phi(\mtX_\textnormal{tr},  \tbTheta, \vch)) 
    + (1 - \alpha) \ccalL_\textnormal{GL} (\mtX_\textnormal{tr},  \mtTheta) \nonumber \\
    &+ \frac{\gamma}{2} \| \mtTheta - \tilde{\mtTheta} \|_F^2 \quad \textnormal{s.t.}  \quad \mtTheta \succeq 0, \quad \| \mtTheta\|_2 \leq M,
\end{align}
where the auxiliary variable $\tbTheta$ decouples $\mathcal{L}_\textnormal{task}$ from $\mathcal{L}_\textnormal{GL}$ through a quadratic penalty. 
Notably, in the limit of $\gamma \to \infty$, the constraint $\mtTheta = \tilde{\mtTheta}$ is enforced exactly, and problem~\eqref{eq:joint_prob} reduces to~\eqref{eq:initial_prob}. 
This alternative formulation is amenable to an alternating optimization strategy~\cite{jin2020graph,tenorio2023robust}.
In particular, we approach it by sequentially solving the following steps for $i=1, \ldots, I_{\text{epochs}}$ iterations.

\vspace{2mm}
\noindent
\textbf{Step 1.}
We update $\mtTheta$ with $\tilde{\mtTheta}^{(i)}$ fixed by solving
\begin{align}\label{eq:step3}
    \mtTheta^{(i+1)} = &\argmin_{\mtTheta} (1 - \alpha) \ccalL_\textnormal{GL} (\mtX_\textnormal{tr},  \mtTheta) + \frac{\gamma}{2} \| \mtTheta - \tilde{\mtTheta}^{(i)} \|_F^2. \\
    &\textnormal{s.t.} \quad \bbTheta \succeq 0, \quad \| \mtTheta\|_2 \leq M. \nonumber
\end{align}
This step amounts to a convex optimization problem closely related to graphical Lasso, which can be solved efficiently with existing algorithms. 
We adopt a proximal gradient descent approach~\cite{beck2017first,navarro2024data} where gradient descent is applied to the smooth terms in \eqref{eq:step3}, followed by the proximal operator of the $\ell_1$ norm (soft-thresholding of the off-diagonal elements), and finally projection onto the cone of positive semidefinite matrices and norm clipping.  

\vspace{2mm}
\noindent
\textbf{Step 2.}
Next, we update $\tbTheta$ while fixing $\vch^{(i)}$ and $\mtTheta^{(i+1)}$. 
This requires solving
\begin{equation}\label{eq:step2}
    \tbTheta^{(i+1)} \!\!\!=\! \argmin_{\tbTheta} \alpha  \mathcal{L}_\textnormal{task}(\vcy_\textnormal{tr}, \!\Phi(\mtX_\textnormal{tr}, \!\tbTheta, \!\vch^{(i)})) \!+\! \frac{\gamma}{2} \! \| \mtTheta^{(i+1)} \!\!-\! \tilde{\mtTheta}  \|_F^2.
\end{equation}
In practice, this step is implemented via a forward pass through the PNN, a computation of the task-related loss and the Frobenius penalty, and gradient descent updates.  

\vspace{2mm}
\noindent
\textbf{Step 3.}
Finally, we update the PNN weights while keeping the previous estimates $\mtTheta^{(i+1)}$ and $\tbTheta^{(i+1)}$ fixed. 
This amounts to solving
\begin{equation}\label{eq:step1}
    \vch^{(i+1)} = \argmin_{\vch} \alpha \mathcal{L}_\textnormal{task}(\vcy_\textnormal{tr}, \Phi(\mtX_\textnormal{tr},  \tbTheta^{(i+1)}, \vch)),
\end{equation}
which corresponds to standard NN training and can be performed with, e.g., the Adam optimizer~\cite{kingma2014adam}. %
We summarize the overall procedure in Algorithm~\ref{alg:alternating} where we let $\mathcal{L}(\vch,\mtTheta,\tilde{\mtTheta})$ denote the objective funtion in~\eqref{eq:joint_prob}. 
Next, we take advantage of the convexity of Step~1 to characterize the quality of the estimated precision matrix. 

\begin{algorithm}[t]
\caption{Iterative optimization of $\mtTheta$, $\tilde{\mtTheta}$, and $\vch$}
\label{alg:alternating}
\begin{algorithmic}[1]
    \State Initialize $\mtTheta^{(1)}$, $\tilde{\mtTheta}^{(1)}$, $\vch^{(1)}$
    \For{$\text{epoch} = 1, \dots, I_{\text{epochs}}$}
        \For{$j = 1, \dots, I_{\Theta}$}
            \State $\mtTheta^{(i+1)}\!\!\! \gets \mtTheta^{(i)}\!\!\! - \eta \left(\mtC -(\mtTheta^{(i)}\!\!+\epsilon \mtI)^{-1}\!\! + \gamma (\mtTheta^{(i)}\!\! - \tilde{\mtTheta}^{(i)})\right)$
            \State $\mtTheta^{(i+1)} \gets \text{soft\_threshold}(\mtTheta^{(i+1)}, \eta\lambda)$
            \State $\mtTheta^{(i+1)} \gets \text{psd\_projection}(\mtTheta^{(i+1)})$
            \State $\mtTheta^{(i+1)} \gets M/\max(M, \|\mtTheta^{(i+1)}\|_2) \mtTheta^{(i+1)}$
        \EndFor
        \For{$k = 1, \dots, I_{\tilde{\Theta}}$}
            \State $\tilde{\mtTheta}^{(i+1)} \gets \tilde{\mtTheta}^{(i)} - \eta \nabla_{\tilde{\mtTheta}} \mathcal{L}(\vch^{(i)},\mtTheta^{(i+1)},\tilde{\mtTheta}^{(i)}))$
        \EndFor
        \For{$l = 1, \dots, I_{h}$}
            \State $\vch^{(i+1)} \gets \text{adam\_update}(\vch^{(i)}, \mathcal{L}(\vch^{(i)},\mtTheta^{(i+1)},\tilde{\mtTheta}^{(i+1)}))$
        \EndFor
    \EndFor
    \State Return $\mtTheta^{(I_\textnormal{epochs}I_\Theta+1)}$, $\tilde{\mtTheta}^{(I_\textnormal{epochs}I_{\tilde{\Theta}}+1)}$, $\vch^{(I_\textnormal{epochs}I_h+1)}$
\end{algorithmic}
\end{algorithm}

\subsection{Theoretical Analysis}
We analyze the error between the sparse precision matrix $\mtTheta^{(i)}$ estimated at iteration $i$ and the true precision matrix $\mtTheta_0$, as a function of the number of samples $T$ and the dimension $N$. 
Our main result is stated in the following theorem. 

\begin{theorem}\label{th:precision_convergence}
    Let $\mtTheta^{(i)}$ denote the minimizer of the objective in~\eqref{eq:step3} with $\lambda \asymp \sqrt{\log N/T}$ at iteration $i$, and let $\mtTheta_0$ be the true precision matrix. 
    Denote by $S$ the number of nonzero off-diagonal entries of $\mtTheta_0$, set $\epsilon = 0$, assume that the eigenvalues of $\mtC_0$ lie in $[k_\textnormal{min}, k_\textnormal{max}]$ for some constants $k_\textnormal{min} > 0$ and $k_\textnormal{max} < \infty$, and assume observations $\vcx_i \sim \mathcal{N}(\mathbf{0}, \mtC_0)$ i.i.d.. 
    Then, with probability tending to one as $N,T \to \infty$, there exist constants $m_1, m_2 > 0$ such that
    \begin{equation}
        \| \mtTheta^{(i)} - \mtTheta_0 \|_F \leq m_1 \sqrt{\frac{(N+S)\log N}{T}} + m_2 \| \mtTheta_0 - \tilde{\mtTheta}^{(i)} \|_*^{1/2}.
    \end{equation}
    where $\|\cdot\|_*$ denotes the nuclear norm.
\end{theorem}
\smallskip
\noindent \textbf{Proof.} The proof is omitted due to space limitations, we refer interested readers to the Appendix of~\cite{cavallo2025pnns}. 

Theorem~\ref{th:precision_convergence} shows that the sequence of estimates $\mtTheta^{(i)}$ obtained from solving~\eqref{eq:step3} converges to the true precision $\mtTheta_0$ as the number of samples increases, with rate $\mathcal{O}(T^{-1/2})$. 
The error bound also depends on the square root of the number of nonzero entries $S$, which makes the method particularly effective when the true precision is highly sparse ($S \ll N^2$). 
In contrast, larger problem dimension $N$ increases the bound, reflecting the inherent difficulty of high-dimensional estimation. 
These dependencies are analogous to those observed for graphical Lasso~\cite{friedman2007lasso}. 
The additional term $\| \mtTheta_0 - \tilde{\mtTheta}^{(i)} \|_*^{1/2}$ captures the distance between the true precision and the task-aware matrix $\tilde{\mtTheta}^{(i)}$, which can be negligible if the task labels reveal meaningful dependency patterns, but may bias the estimate away from the true precision if such patterns are not aligned with the underlying statistical dependencies. This result justifies the role of the decopling strategy that aims at learning jointly the precision matrix and the PNN weights while remaining principled to the ground truth precision. 

\begin{figure*}[t]
    \centering
    \includegraphics[width=\linewidth]{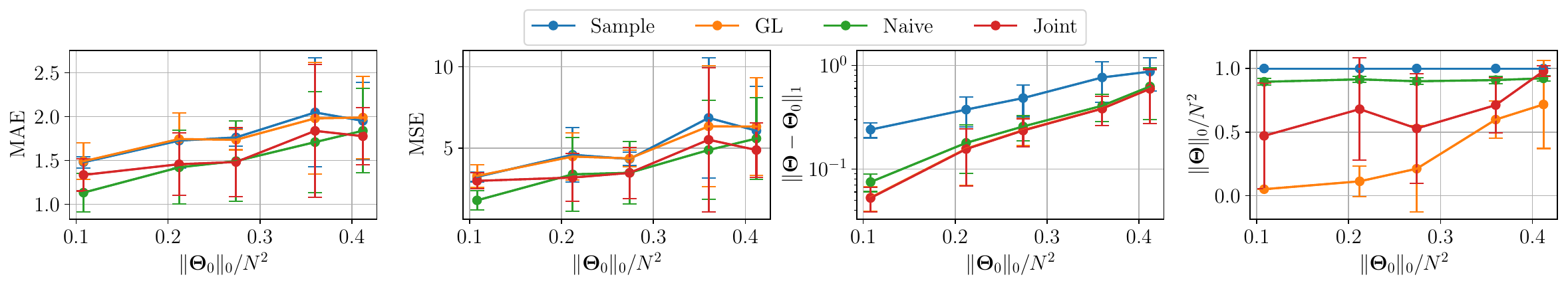}
    \caption{Regression error (MAE and MSE), precision matrix reconstruction quality in terms of L1 norm and sparsity of estimated precision $\mtTheta$ on the synthetic datasets for varying sparsity of the true precision $\mtTheta_0$.\vspace{-.4cm}}
    \label{fig:synth_data}
\end{figure*}

\section{Numerical Results}
We evaluate PNNs over synthetic and real-world datasets with two main objectives: \textbf{(O1)} to validate the superior downstream performance of joint optimization compared to two-stage approaches, and \textbf{(O2)} to assess the ability of our method to produce informative and sparse precision estimates.

\subsection{Baselines and Hyperparameters}
We compare four variants of PNN: PNN on the sample precision (\textbf{Sample}); PNN on the precision estimated via graphical Lasso (\textbf{GL}); PNN with joint precision estimation using the naive formulation in~\eqref{eq:initial_prob} followed by a final soft-thresholding step (\textbf{Naive}); and PNN trained with the alternating optimization in Algorithm~\ref{alg:alternating} (\textbf{Joint}). 
On real data, we compare with two additional baselines: \textbf{VNN}~\cite{sihag2022covariance} and PCA projection followed by an MLP regressor (\textbf{PCA}).
For all models and datasets, hyperparameters are tuned on a validation set over the following ranges: number of layers $L \in \{1,2,3\}$, filterbank size $F \in \{8,16\}$, filter order $K \in \{1,2,3\}$, $\lambda_0 \in \{1,10,20\}$ where $\lambda=\lambda_0 \sqrt{\log N/T}$, $\alpha=0.5$, learning rate $\eta = 0.01$, $\gamma = 10$, $I_\textnormal{epochs} = 10$, and $I_\Theta = I_{\tilde{\Theta}} = I_h = 20$. 
Batch normalization is applied to all PNNs. 
For synthetic data, hyperparameters are selected on the setting with sparsity $0.2$ and reused for the other settings. 
All datasets are split into train/validation/test sets of size $60\%/20\%/20\%$, and each experiment is repeated 5 times, reporting mean and standard deviation.
All code and additional implementation details are available on GitHub\footnote{\scriptsize \url{https://github.com/andrea-cavallo-98/PNN}}.

\subsection{Synthetic Data}
We generate synthetic datasets by defining a precision matrix $\mtTheta_0$ with prescribed sparsity ($\|\mtTheta_0\|_0 / N^2$) and sampling observations as $\vcx_i \sim \mathcal{N}(\mathbf{0}, \mtTheta_0^{-1})$. 
We stack the samples into a matrix $\mtX \in \mathbb{R}^{N\times T}$ and create regression targets as $\vcy = \vcw^\Tr \mtX + \vcz$, with $\vcw \sim \mathcal{N}(\mathbf{0}, \mtI)$ and noise entries $z_i \sim \mathcal{N}(0,\sigma)$, where $\sigma$ is set to achieve an SNR of 10. 
We set $N = 20$ and $T = 100$. 
From Figure~\ref{fig:synth_data}, \textbf{Joint} and \textbf{Naive} achieve comparable or better performance than \textbf{GL} and \textbf{Sample} across all sparsity levels, in terms of both MAE and MSE. 
\textbf{Joint} consistently provides the best results, confirming the benefit of the joint optimization for downstream performance \textbf{(O1)}. 
Regarding precision reconstruction quality, \textbf{GL} and \textbf{Joint} perform similarly in terms of $\ell_1$ error, while \textbf{Naive} yields poorer estimates and \textbf{Sample} is particularly inaccurate due to the numerous spurious entries \textbf{(O2)}. 
Finally, both \textbf{GL} and \textbf{Joint} recover sparsity levels close to the ground truth, enabling efficient implementations. 
In contrast, \textbf{Sample} contains no zero entries due to noise, and \textbf{Naive} produces very few zeros as a result of its unstable training dynamics.

\subsection{Real Data}
We further evaluate PNNs on two brain datasets, ADNI and ABIDE, containing cortical thickness measurements derived from MRI scans. 
ADNI~\cite{jack2008alzheimer} includes patients with Alzheimer’s disease and healthy controls. 
We use the ADNI2 collection, extracting cortical thickness measures from the FreeSurfer outputs~\cite{fischl2012freesurfer} available at \footnote{\scriptsize \url{https://ida.loni.usc.edu/}}, resulting in $N = 68$ brain regions and $T = 1142$ patients. 
ABIDE~\cite{craddock2013neuro} includes patients with autism and healthy controls. 
We obtain cortical thickness measures following the steps described at \footnote{\scriptsize \url{http://preprocessed-connectomes-project.org/abide/}}, retaining the features of the Mindboggle protocol~\cite{klein2017mindboggling}, which yields $N = 62$ and $T = 1035$. 
Our goal is to predict patient age from brain measurements, a problem of interest in neuroscience since discrepancies between brain age and chronological age are indicative of neurodegenerative disease~\cite{sihag2024explainable}. 

Table~\ref{tab:real_data} reports the results. 
Overall, \textbf{Joint} achieves the best performance in terms of MAE on the age prediction task \textbf{(O1)}, while also producing sparse precision estimates \textbf{(O2)}, confirming the benefit of the joint estimation for real-world downstream tasks. 
Both \textbf{GL} and \textbf{Joint} outperform \textbf{Sample} by relying on cleaner precision estimates, further corroborating that sample estimates in real data contain noisy dependencies that hinder predictive performance. \textbf{VNN} performs closely to \textbf{GL} on ABIDE, but worse on ADNI, as it also operates on the sample covariance which contains spurious correlations. \textbf{PCA}, finally, performs well on ABIDE but particularly badly on ADNI, indicating that more complex covariance processing is helpful to achieve consistent performance across settings.

\begin{table}[t]
\footnotesize
\centering
\begin{tabular}{c|cc|cc}
\toprule
\multirow{2}{*}{\textbf{Method}} & \multicolumn{2}{c|}{\textbf{ADNI}} & \multicolumn{2}{c}{\textbf{ABIDE}} \\
 & MAE & \#Zeros & MAE & \#Zeros \\
\midrule
PCA & 10.3$\pm$0.7 & -- & 4.41$\pm$0.34 & -- \\
VNN & 6.06$\pm$0.20 & -- & 5.81$\pm$0.11 & -- \\
Sample & 5.78$\pm$0.10 & 0$\pm$0 & 5.94$\pm$0.02 & 0$\pm$0 \\
GL     & 5.81$\pm$0.06 & 4546$\pm$0 & 5.80$\pm$0.11 & 2232$\pm$0 \\
Joint  & \textbf{5.31$\pm$0.18} & 1225$\pm$1126 & \textbf{3.64$\pm$0.65} & 2771$\pm$976 \\
\bottomrule
\end{tabular}
\caption{Regression error (MAE) and number of zero elements in the estimated precision matrix for the age prediction task from cortical thickness measurements. We do not report sparsity metrics for PCA and VNN as they do not use the precision matrix.\vspace{-.4cm}}
\label{tab:real_data}
\end{table}

\vspace{-.3cm}
\section{Conclusion}
We introduced PNNs, a class of graph convolutional NNs that leverage sparse estimates of the precision matrix, and we proposed a novel paradigm based on alternating optimization to jointly estimate the NN weights and the sparse precision matrix.
We established theoretical guarantees on the stability of the estimated sparse precision with respect to the true precision, considering both the number of available observations and the downstream task.
Our numerical experiments on synthetic and real datasets, focusing on age prediction from cortical thickness measurements, demonstrated the effectiveness of the proposed joint learning strategy.
Future work will focus on improving the interpretability of the task-aware precision matrices, exploring the stability of PNNs in low-data regimes, and further analyzing the convergence properties of the joint optimization.

\bibliographystyle{IEEEbib}
\bibliography{myIEEEabrv,bibliography}

\newpage

\appendix

\section{Proof of Theorem 1}

We provide the proof of Theorem 1, which extends those in~\cite[Theorem 1]{rothman2008sparse} and~\cite[Theorem 1]{navarro2024fair}.
Let $\mathcal{L}_\mtTheta(\mtTheta)$ denote the loss in~\eqref{eq:step3} for a fixed $\tilde{\mtTheta}$.
Consider the function $Q(\mtTheta) = \mathcal{L}_\mtTheta(\mtTheta) - \mathcal{L}_\mtTheta(\mtTheta_0)$ measuring the difference between the loss of a generic matrix $\mtTheta$ and that of the true precision $\mtTheta_0$. By definition of $\mtTheta^{(i)}$ as the minimizer of $\mathcal{L}_\mtTheta(\mtTheta)$, we have that $Q(\mtTheta^{(i)}) \leq 0$. 
If we can find a bound $B$ such that $\| \mtTheta - \mtTheta_0 \|_F > B$ implies $Q(\mtTheta)>0 $, then we have that $Q(\mtTheta) \leq 0$ implies $\| \mtTheta - \mtTheta_0 \|_F \leq B$ and we can conclude that $\| \mtTheta^{(i)} - \mtTheta_0 \|_F \leq B$.
In the following, we compute a value for $B$.

Let us now expand the function $Q(\mtTheta)$ and rearrange its terms:
\begin{align}
    Q(\mtTheta) = (1-\alpha)(-\operatorname{logdet}(\mtTheta) + \operatorname{tr}(\mtC \mtTheta) + \lambda \| \mtTheta_\mathcal{\bar{D}} \|_1) + \nonumber \\ \gamma/2 \| \mtTheta - \tilde{\mtTheta} \|_F^2 
    +(1-\alpha)(\operatorname{logdet}(\mtTheta_0) - \operatorname{tr}(\mtC \mtTheta_0) - \nonumber\\ \lambda \| (\mtTheta_0)_\mathcal{\bar{D}} \|_1) - \gamma/2 \| \mtTheta_0 + \tilde{\mtTheta} \|_F^2 \nonumber\\
    = (1-\alpha)(\underbrace{\operatorname{tr}((\mtC - \mtC_0)(\mtTheta - \mtTheta_0))}_{\textnormal{Term 1}} + \nonumber\\ \underbrace{\operatorname{tr}(\mtC_0(\mtTheta - \mtTheta_0)) - (\operatorname{logdet}\mtTheta-\operatorname{logdet}\mtTheta_0)}_{\textnormal{Term 2}} + \nonumber\\\underbrace{\lambda(\| \mtTheta_\mathcal{\bar{D}} \|_1 - \| (\mtTheta_0)_\mathcal{\bar{D}}  \|_1)}_{\textnormal{Term 3}} )+ \underbrace{\frac{\gamma}{2}(\| \mtTheta - \tilde{\mtTheta} \|_F^2 - \| \mtTheta_0 - \tilde{\mtTheta} \|_F^2)}_{\textnormal{Term 4}}.\nonumber
\end{align}
We proceed to bound Terms 1-4. Let $r_T = \sqrt{\frac{(N+S)\log N}{T}}$.

\smallskip
\noindent \textbf{Term 1.}
For Term 1, we use the result from~\cite[(22)]{navarro2024fair}: there exist some constants $c_1,c_2$ such that, with probability tending to 1 as $N \rightarrow \infty$, the following holds.

\begin{align}
    \operatorname{tr}((\mtC - \mtC_0)(\mtTheta - \mtTheta_0)) \geq& -c_1 \sqrt{\frac{\log N}{T}} \| \mtTheta_\mathcal{\bar{D}} - (\mtTheta_0)_\mathcal{\bar{D}} \|_1 \nonumber \\
    & -c_2 r_T \| \mtTheta_\mathcal{D} - (\mtTheta_0)_\mathcal{D} \|_F \label{eq:t1}
\end{align}
where $\mtTheta_\mathcal{D}$ and $\mtTheta_\mathcal{\bar{D}}$ are matrixes containing the diagonal and off-diagonal values of $\mtTheta$, respectively.

\smallskip
\noindent \textbf{Term 2.}
For Term 2, we leverage the result from~\cite[(29)]{navarro2024fair}:
\begin{align}
    \operatorname{logdet}\mtTheta-\operatorname{logdet}\mtTheta_0 \leq & \operatorname{tr}(\mtC(\mtTheta - \mtTheta_0)) \nonumber \\
    & - \frac{1}{2\tau} k_\textnormal{min}^2\| \mtTheta - \mtTheta_0 \|^2_F \label{eq:t2}
\end{align}
where $\tau = \max\{ 4, (1+k_\textnormal{min}\|\mtTheta - \mtTheta_0\|_F^2) \}$.

\smallskip
\noindent \textbf{Term 3.}
For Term 3, we use the result in \cite[(30)]{navarro2024fair}:
\begin{align}
    \lambda(\| \mtTheta_\mathcal{\bar{D}} \|_1 - \| (\mtTheta_0)_\mathcal{\bar{D}}  \|_1) \geq & \lambda ( \| (\mtTheta - \mtTheta_0)_{\mathcal{\bar{D}} \cap \mathcal{\bar{S}}} \|_1 - \nonumber\\
    & \| (\mtTheta - \mtTheta_0)_{\mathcal{\bar{D}} \cap \mathcal{S}} \|_1 )  \label{eq:t3}
\end{align}
where $\mathcal{S}$ selects the elements of a matrix corresponding to the non-zero off-diagonal entries of $\mtTheta_0$ and $\mathcal{\bar{S}}$ the complementary entries.

\smallskip
\noindent \textbf{Term 4.}
We provide a lower bound for Term 4 by exploiting the property that $H(\mtA+\mtB) - H(\mtA) \geq \operatorname{tr}(\nabla H(\mtA)^\Tr \mtB)$ for any function $H$ differentiable and convex in $\mtA$ and the property that $\operatorname{tr}(\mtA\mtB) \geq -\| \mtA \|_* \|\mtB\|_2$. If we set $H(\mtA) = \| \mtA - \tilde{\mtTheta} \|_F^2$, then $\nabla H(\mtA) = 2(\mtA - \tilde{\mtTheta})$ and term 4 can be bounded as 
\begin{align}
    \frac{\gamma}{2}(\| \mtTheta - \tilde{\mtTheta} \|_F^2 - \| \mtTheta_0 - \tilde{\mtTheta} \|_F^2) \geq \nonumber\\
    \gamma \operatorname{tr}((\mtTheta_0 - \tilde{\mtTheta})(\mtTheta_0 - \mtTheta)) \geq \nonumber \\
    -\gamma \| \mtTheta_0 - \tilde{\mtTheta} \|_* (\| \mtTheta_0 \|_2 + \| \mtTheta \|_2). \label{eq:t4}
\end{align}

\smallskip
\noindent \textbf{Final bound.}
We now combine \eqref{eq:t1}, \eqref{eq:t2}, \eqref{eq:t3}, \eqref{eq:t4} to provide a complete lower bound for $Q(\mtTheta)$.
\begin{align}
    Q(\mtTheta) \geq& (1-\alpha)(-c_1 \sqrt{\frac{\log N}{T}} \| \mtTheta_\mathcal{\bar{D}} - (\mtTheta_0)_\mathcal{\bar{D}} \|_1 - \nonumber \\ & c_2 r_T \| \mtTheta_\mathcal{D} - (\mtTheta_0)_\mathcal{D} \|_F +\frac{1}{2\tau} k_\textnormal{min}^2\| \mtTheta - \mtTheta_0 \|^2_F + \nonumber \\
     & \lambda ( \| (\mtTheta - \mtTheta_0)_{\mathcal{\bar{D}} \cap \mathcal{\bar{S}}} \|_1 - \| (\mtTheta - \mtTheta_0)_{\mathcal{\bar{D}} \cap \mathcal{S}} \|_1 ) ) - \nonumber \\
     & \gamma \| \mtTheta_0 - \tilde{\mtTheta} \|_* (\| \mtTheta_0 \|_2 + \| \mtTheta \|_2).
\end{align}
Now, let us set $\lambda = c_1/\epsilon_1 \sqrt{\log N/T}$ for $\epsilon_1 < 1$ (see Theorem's assumptions) and let us introduce a constant $c_3 \geq \| \mtTheta_0\|_2 + \|\mtTheta\|_2$ (since $\|\mtTheta\|_2$ is bounded by the problem formulation).
By rearranging terms and leveraging the fact that $\| (\mtTheta - \mtTheta_0)_\mathcal{D} \|^2_F \leq \| \mtTheta - \mtTheta_0 \|^2_F$ and $\| (\mtTheta - \mtTheta_0)_{\bar{\mathcal{D}}} \|^2_F \leq \| \mtTheta - \mtTheta_0 \|^2_F$ by properties of the Frobenius norm, we obtain:
\begin{align}
    Q(\mtTheta) \geq & (1-\alpha)( - c_1 (1+1/\epsilon_1) r_T\| (\mtTheta - \mtTheta_0)_\mathcal{\bar{D}} \|_F +  \nonumber \\
    & \frac{1}{2\tau} k_\textnormal{min}^2\| \mtTheta- \mtTheta_0 \|^2_F - c_2 r_T\| (\mtTheta - \mtTheta_0)_\mathcal{D} \|_F) \nonumber \\
    & - c_3 \gamma \| \mtTheta_0 - \tilde{\mtTheta} \|_* \geq \nonumber \\
    & (1-\alpha)( - c_1 (1+1/\epsilon_1) r_T\| \mtTheta - \mtTheta_0 \|_F +  \nonumber \\
    & \frac{1}{2\tau} k_\textnormal{min}^2\| \mtTheta- \mtTheta_0 \|^2_F - c_2 r_T\| \mtTheta - \mtTheta_0 \|_F) \nonumber \\
    & - c_3 \gamma \| \mtTheta_0 - \tilde{\mtTheta} \|_* = \nonumber \\
    & (1-\alpha) \| \mtTheta - \mtTheta_0 \|_F \Bigg( \frac{k_\textnormal{min}^2}{4\tau} \| \mtTheta - \mtTheta_0 \|_F  \nonumber \\
    & - c_1(1+\frac{1}{\epsilon_1})r_T -c_2r_T \Bigg) \label{eq:s1} \\
    & + (1-\alpha)\frac{k_\textnormal{min}^2}{4\tau} \| \mtTheta - \mtTheta_0 \|_F^2 - c_3\gamma \| \mtTheta_0 - \tilde{\mtTheta} \|_* \label{eq:s2}.
\end{align}
We now derive conditions on $\| \mtTheta - \mtTheta_0 \|_F$ from \eqref{eq:s1}, \eqref{eq:s2}, such that $Q(\mtTheta)$ is positive. From \eqref{eq:s1}, we have that 
\begin{equation}\label{eq:c1}
    \| \mtTheta - \mtTheta_0 \|_F > 4\tau k_\textnormal{min}^{-2} \left(c_1 \left(1+\frac{1}{\epsilon_1}\right) r_T - c_2 r_T\right).
\end{equation}
From \eqref{eq:s2}, we get 
\begin{equation}\label{eq:c2}
    \| \mtTheta - \mtTheta_0 \|_F^2 > 4\tau k_\textnormal{min}^{-2}c_3\gamma \| \mtTheta_0 - \tilde{\mtTheta} \|_*.
\end{equation}
If both conditions are satisfied, then $Q(\mtTheta)>0$. 
A sufficient condition for this is to require $\| \mtTheta - \mtTheta_0 \|_F$ to be larger than the maximum between the bounds in \eqref{eq:c1} and \eqref{eq:c2}.
By rearranging such terms, it follows that there exist $m_1, m_2$ such that, with probability tending to 1 as $N,T\rightarrow \infty$, the condition
\begin{align}
    \| \mtTheta - \mtTheta_0 \|_F > m_1 r_T + m_2 \| \mtTheta_0 - \tilde{\mtTheta} \|_*^{1/2}
\end{align}
implies that $Q(\mtTheta) > 0$ and this concludes the proof. \hfill $\blacksquare$

\end{document}